\documentclass{article} %
\usepackage[preprint]{colm2026_conference}

\usepackage{microtype}
\usepackage{hyperref}
\usepackage{url}
\usepackage{booktabs}
\usepackage{graphicx}
\usepackage{amsmath}
\usepackage{multirow}

\usepackage{booktabs}
\usepackage{graphicx}
\usepackage{xcolor}
\usepackage{tikz}
\usepackage{array}
\usepackage{colortbl}
\usepackage{microtype}  %
\usepackage{subfigure}
\usepackage{caption}
\usepackage{wrapfig}
\usepackage{siunitx}
\usepackage[T1]{fontenc}
\usepackage{fvextra}
\usepackage{tcolorbox}
\tcbuselibrary{breakable, skins}

\definecolor{orangebg}{RGB}{253, 236, 231}
\definecolor{closedorange}{RGB}{216, 90, 48}
\definecolor{closedorangelight}{RGB}{232, 160, 96}
\definecolor{openteal}{RGB}{15, 110, 86}
\definecolor{openteallight}{RGB}{29, 158, 117}
\definecolor{grouporange}{RGB}{196, 98, 45}
\definecolor{groupteal}{RGB}{15, 110, 86}
\definecolor{barbg}{RGB}{220, 218, 214}
\definecolor{groupline}{RGB}{200, 198, 194}

\newcommand{\pbarrounded}[3]{%
  \begin{tikzpicture}[baseline=-0.4ex]
    \filldraw[barbg, rounded corners=1.4pt]
      (0,0) rectangle (2.8cm, 0.14cm);
    \begin{scope}
      \clip[rounded corners=1.4pt] (0,0) rectangle (2.8cm,0.14cm);
      \shade[left color=#2, right color=#3]
        (0,0) rectangle (#1 * 0.028cm, 0.14cm);
    \end{scope}
  \end{tikzpicture}%
}

\newcommand{\mlogo}[1]{%
  \raisebox{-1.8pt}{\includegraphics[height=11pt]{figs/logo/#1}}%
}

\newcommand{\grouphdr}[2]{%
  {\fontsize{7.5}{9}\selectfont\color{#2}%
   \textls[80]{\MakeUppercase{#1}}}%
}

\newcommand{\bestclosed}[1]{{\color{closedorange}\textbf{#1}}}
\newcommand{\bestopen}[1]{{\color{openteal}\textbf{#1}}}

\usepackage{lineno}

\definecolor{darkblue}{rgb}{0, 0, 0.5}
\hypersetup{colorlinks=true, citecolor=darkblue, linkcolor=darkblue, urlcolor=darkblue}
\definecolor{sqlboxbg}{RGB}{248,248,246}
\definecolor{sqlboxframe}{RGB}{218,218,214}
\definecolor{sqlkw}{RGB}{32,96,160}
\definecolor{sqlfn}{RGB}{158,82,40}
\definecolor{sqlstr}{RGB}{15,110,86}
\newcommand{\aifunc}[1]{\textbf{\textcolor{sqlfn}{#1}}}

\newtcolorbox{sqlexamplebox}{
  enhanced,
  colback=sqlboxbg,
  colframe=sqlboxframe,
  boxrule=0.35pt,
  arc=3pt,
  width=0.82\linewidth,
  center,
  left=7pt,
  right=7pt,
  top=5pt,
  bottom=5pt,
  before skip=0.55em,
  after skip=0.65em,
  boxsep=0pt
}

\newcommand{\resourcelink}[3]{%
  \href{#1}{%
    \tcbox[
      enhanced,
      on line,
      colback=orangebg!45,
      colframe=grouporange!45,
      boxrule=0.45pt,
      arc=4pt,
      left=5pt,
      right=6pt,
      top=2.3pt,
      bottom=2.3pt,
      boxsep=0pt
    ]{\raisebox{-1.8pt}{\includegraphics[height=9pt]{figs/logo/#2}}\hspace{3pt}{\footnotesize\sffamily\fontseries{bx}\selectfont\textcolor{grouporange}{#3\,{\scriptsize$\nearrow$}}}}%
  }%
}

\newcommand{\resourcelinks}{%
  \begin{center}
    \vspace{-0.55em}
    \resourcelink{https://github.com/Leolty/Spider2-AIFunc}{github.png}{GitHub}%
    \hspace{0.5em}%
    \resourcelink{https://huggingface.co/datasets/tianyang/spider2-aifunc}{huggingface.png}{HuggingFace}%
    \vspace{-0.45em}
  \end{center}%
}

\title{Spider 2.0-AIFunc: Extending Real-World Text-to-SQL to AI-Native SQL Workflows}

\author{%
Tianyang Liu$^{1}$\thanks{Work done during Tianyang Liu's internship at Snowflake.},\hspace{0.35em}
Canwen Xu$^{2}$,\hspace{0.35em}
Fangyu Lei$^{3}$,\hspace{0.35em}
Nikki Lijing Kuang$^{2}$, \\
\textbf{Jixuan Chen}$^{1}$,\hspace{0.35em}
\textbf{Tao Yu}$^{3}$,\hspace{0.35em}
\textbf{Julian McAuley}$^{1}$,\hspace{0.35em}
\textbf{Zhewei Yao}$^{2}$,\hspace{0.35em}
\textbf{Yuxiong He}$^{2}$ \\
$^{1}$UC San Diego \quad
$^{2}$Snowflake AI Research \quad
$^{3}$University of Hong Kong \\
\texttt{\{til040,jmcauley\}@ucsd.edu} \quad
\texttt{\{canwen.xu,zhewei.yao,yuxiong.he\}@snowflake.com}
}

\begin{document}

\ifcolmsubmission
\linenumbers
\fi

\maketitle

\vspace{-0.6cm}

\resourcelinks

\begin{abstract}
Major cloud data platforms now expose large language model capabilities
as native SQL functions, enabling analysts to perform classification,
filtering, sentiment analysis, extraction, similarity search, and
aggregation within ordinary SQL queries. Yet existing
text-to-SQL benchmarks evaluate only conventional SQL and provide no
signal on whether models can generate such AI-native SQL. We introduce
Spider 2.0-AIFunc, a benchmark of 465 verified instances across 125
real-world databases covering six types of AI functions on the Snowflake
platform. Starting from an existing enterprise text-to-SQL benchmark, we
construct Spider 2.0-AIFunc through an agent-based pipeline that rewrites
source tasks into AI-native form, simultaneously transforming target
queries and refining natural language instructions to make the
intended AI-native solution explicit and reduce ambiguity. All instances pass a
multi-round repeated execution protocol across temporally separated
windows to confirm result stability before release. Evaluating ten
state-of-the-art language models, we find that the strongest proprietary
models reach 67--70\% execution accuracy while the best open-source
model achieves 58.1\%, a gap driven primarily by errors in predicate
specification, schema grounding, and AI function parameterization.
Agent frameworks designed for traditional text-to-SQL challenges, such
as schema retrieval and relevant table selection, do not transfer
effectively to AI-native SQL: a minimal agent setup consistently matches
or outperforms more elaborate alternatives, suggesting that the
strategies these frameworks employ are less critical in this setting.
\end{abstract}

\vspace{-0.3cm}

\section{Introduction}

Much of enterprise data analysis begins with questions that are
easy to ask in natural language but awkward to express in
traditional SQL: Which support tickets describe billing issues?
Which survey responses mention churn risk? Which product reviews
are negative, and what are customers complaining about? Consider
the last case. A data analyst has a table of product reviews stored
in a cloud data warehouse and wants to analyze their sentiment and
categorize the negative ones by complaint type. Until recently,
answering this kind of question meant extracting the text from the
database, running sentiment analysis and classification models in a
separate environment, and loading the results back. Today, on
platforms like Snowflake, this can be done directly in SQL using
AI functions such as \texttt{AI\_SENTIMENT} and
\texttt{AI\_CLASSIFY}:

\begin{sqlexamplebox}
\begin{Verbatim}[
fontsize=\footnotesize,
breaklines=true,
commandchars=\\\{\}
]
\textcolor{sqlkw}{SELECT}
  review_id,
  \aifunc{AI\_SENTIMENT}(review_text) \textcolor{sqlkw}{AS} sentiment,
  \aifunc{AI\_CLASSIFY}(
    review_text,[\textcolor{sqlstr}{'shipping'}, \textcolor{sqlstr}{'quality'}, \textcolor{sqlstr}{'service'}, \textcolor{sqlstr}{'pricing'}]
  ):labels \textcolor{sqlkw}{AS} complaint_type
\textcolor{sqlkw}{FROM} product_reviews;
\end{Verbatim}
\end{sqlexamplebox}

Such functions can be combined with standard SQL aggregations,
filters, and joins, enabling analysts to express semantic operations
over unstructured text within ordinary SQL queries~\citep{snowflake_cortex_ai_functions}.
Snowflake is not alone in this direction. BigQuery~\citep{bigquery_ai_intro},
Databricks~\citep{databricks_ai_functions}, and other major cloud
platforms have all introduced SQL-callable AI functions, enabling a
broad spectrum of semantic operations, from text classification and
sentiment analysis to similarity search and information extraction,
to be expressed directly within SQL queries. In enterprise settings,
analysts increasingly compose these AI functions with conventional SQL
operators to build AI-native SQL workflows that would have previously
required separate toolchains.

\begin{figure}
    \centering
    \includegraphics[width=\linewidth]{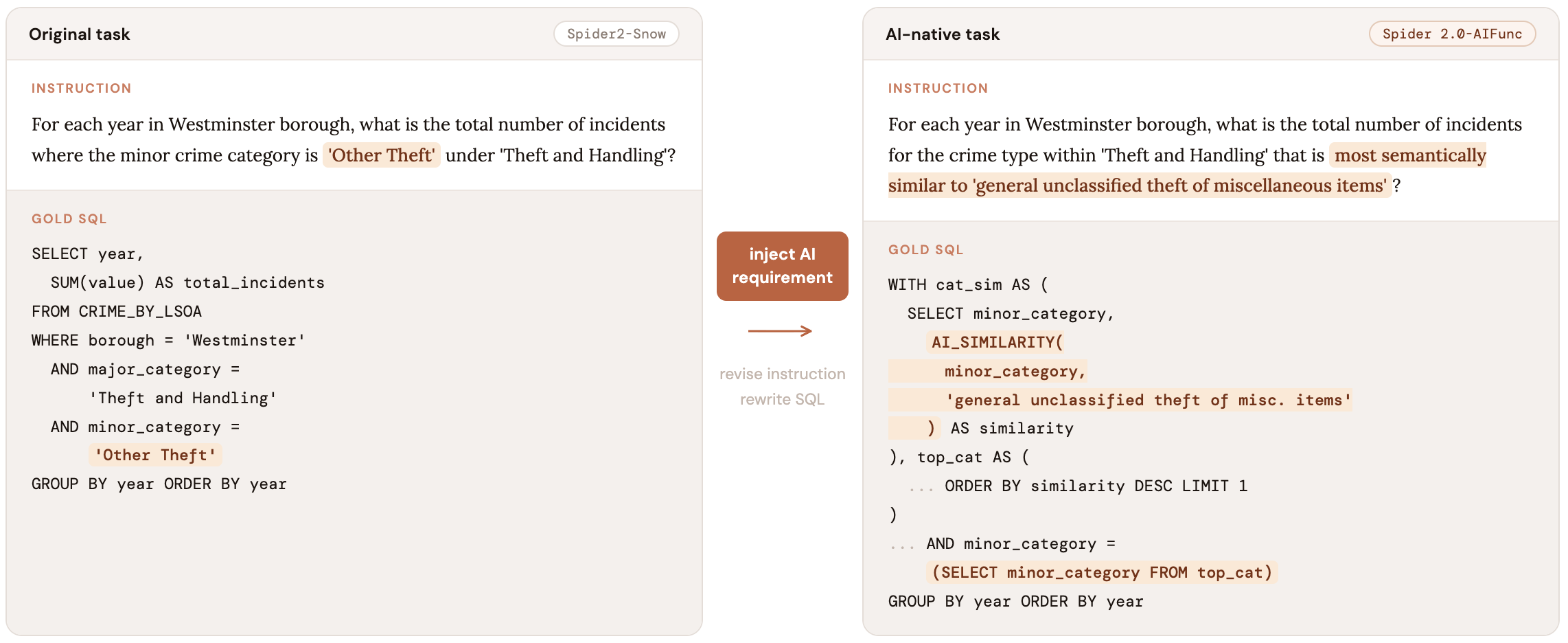}
    \caption{An example of transforming a traditional SQL task into an AI-native SQL instance in \textsc{Spider~2.0-AIFunc}. The original
instruction and gold SQL are rewritten to incorporate a Snowflake
Cortex AI function (\texttt{AI\_SIMILARITY}). Highlighted spans
indicate the injected AI requirement and the corresponding SQL
transformation.}
    \label{fig:example}
    \vspace{-0.6cm}
\end{figure}

Text-to-SQL benchmarks have steadily evolved toward more
realistic evaluation settings~\citep{liu2025surveytexttosqlerallms}. WikiSQL~\citep{zhong2017seq2sqlgeneratingstructuredqueries} introduced
large-scale evaluation over simple single-table queries. Spider
1.0~\citep{yu2019spiderlargescalehumanlabeleddataset} raised the complexity with cross-database
generalization, joins, and nested subqueries. BIRD~\citep{li2023llmservedatabaseinterface}
brought in real-world database values and external knowledge. Most
recently, Spider 2.0~\citep{lei2025spider20evaluatinglanguage} and BEAVER~\citep{chen2025beaverenterprisebenchmarktexttosql} moved to enterprise-level
workflows with multi-dialect SQL over databases that often exceed
1,000 columns. These benchmarks have collectively driven
significant progress in text-to-SQL research. At the same time,
they all target SQL queries composed entirely of conventional SQL
operators over structured data. Given the growing adoption of AI
functions in enterprise SQL workflows, as described above, a
natural next step is to extend text-to-SQL evaluation to cover
AI-native SQL, where the expected output includes AI function
calls alongside traditional SQL operations. Figure~\ref{fig:example}
shows a representative example of this transformation.

To this end, Spider 2.0-AIFunc extends Spider2-Snow~\citep{lei2025spider20evaluatinglanguage} by rewriting its source tasks into AI-native SQL versions.
Starting from 513 Spider2-Snow instances, each with its original
instruction, one or more gold SQL queries, their execution
results, the database schema, and optional external knowledge,
we simultaneously transform the target SQL to incorporate
Snowflake Cortex AI Functions and revise the corresponding
instruction to match. When a source instance has multiple valid
gold SQLs, we further refine the instruction to reduce ambiguity
and make the expected AI-native SQL query more explicitly
specified. We design a reusable, agent-based
pipeline to carry out this construction: agents interact directly
with the Snowflake environment, proposing modifications to both
the SQL and the instruction, executing the resulting query against
the database, and iteratively resolving execution errors,
timeouts, and malformed outputs until the query returns reasonable
results within an acceptable time frame. The agents also check
that AI function parameters are explicitly specified (e.g., the
complete set of class labels for \texttt{AI\_CLASSIFY}) and that
the instruction clearly indicates the expected AI function usage.
After construction, each instance goes through a
verification stage where a separate agent executes the gold SQL
multiple times across several passes, repairing instances that
produce inconsistent results. The benchmark is
constructed in two rounds: a main round and a diversity round
that broadens coverage across AI function types. The released
benchmark comprises 465 verified instances across 125 databases, covering six types of Snowflake Cortex AI Functions.

We evaluate a range of state-of-the-art language models on
Spider 2.0-AIFunc using the Spider-Agent framework~\citep{lei2025spider20evaluatinglanguage},
where each model is provided with a minimal tool set consisting of bash
for exploring database schemas and external knowledge, a SQL execution
tool for running queries against Snowflake, and a task termination tool.
Models are additionally given reference documentation for the relevant
Snowflake Cortex AI Functions. Among the models we evaluate, Claude
Opus 4.6 achieves the highest execution accuracy at 70.3\%, followed
by Claude Sonnet 4.6 at 69.0\% and Gemini 3.1 Pro at 67.1\%. A
consistent gap emerges between proprietary and open-source models: the
strongest proprietary models cluster in the 67--70\% range, while the
best open-source model reaches 58.1\% and the remaining open-source
models range from 44.9\% to 57.0\%. Agent frameworks specifically
designed for traditional text-to-SQL challenges do not transfer
effectively to AI-native SQL: three such frameworks, when given the same
backbone model and AI function documentation, match but do not exceed
the minimal Spider-Agent baseline.

\section{Background and Preliminaries}

\subsection{Spider2-Snow}

Spider 2.0~\citep{lei2025spider20evaluatinglanguage} is an enterprise-level text-to-SQL
benchmark comprising 632 real-world workflow problems over
databases sourced from industrial applications such as Google
Analytics and Salesforce, hosted across multiple database systems
including BigQuery, Snowflake, SQLite, DuckDB, and PostgreSQL.
It offers several evaluation settings of varying complexity.
Spider 2.0-AIFunc builds on \textbf{Spider2-Snow}, a self-contained text-to-SQL setting in which all 547
examples are hosted entirely on Snowflake. In Spider2-Snow,
given a natural language question $Q$, a database schema $D$,
and auxiliary documentation $E$, a text-to-SQL parser $f(\cdot)$
is required to produce the corresponding SQL query
$S = f(Q, D, E \mid \theta)$. Performance is measured by checking whether the predicted SQL
produces the same result as the gold SQL when executed against
the database. A notable aspect of Spider 2.0's evolution is
that each task was originally released with a single gold SQL
query; however, as more teams submitted results and explored
alternative solutions, the maintainers discovered that many
tasks admit multiple valid SQL queries that produce different correct
results, leading to ongoing updates to the evaluation suite.

\subsection{Snowflake Cortex AI Functions}
\label{sec:aifunc}

Snowflake Cortex AI Functions~\citep{snowflake_cortex_ai_functions} are a family of
built-in SQL functions provided by the Snowflake platform that
invoke large language models as part of query execution. Users
call these functions in SQL statements just like conventional
SQL functions, but instead of performing deterministic
operations such as arithmetic or string manipulation, they
perform semantic operations such as classification, sentiment
analysis, information extraction, and similarity computation.
A typical AI function takes one or more table columns (usually
containing text) along with user-specified parameters as input,
and returns its output as part of the query result. Because they
conform to standard SQL syntax, AI functions can be freely
composed with traditional SQL operators such as \texttt{WHERE},
\texttt{JOIN}, \texttt{GROUP BY}, and aggregation functions
within a single query.

Under the hood, each AI function call is served by a language
model available in the Snowflake environment. At any given
point in time, the underlying model for each function is fixed,
and inference is run with a temperature of zero, meaning that
the same input should in principle produce the same output. In
practice, however, prior work has shown that LLM outputs can
exhibit minor inconsistencies even at temperature zero~\citep{he2025nondeterminism, atil2025nondeterminismdeterministicllmsettings, Ouyang_2025}. This property is
relevant to benchmarking because evaluation via execution
accuracy requires that the gold SQL produces a stable result.
We describe how we address this through multi-round execution
verification in Section~\ref{sec:benchmark}.

\section{The Spider 2.0-AIFunc Benchmark}
\label{sec:benchmark}

Spider 2.0-AIFunc is built on top of Spider2-Snow by rewriting
its source tasks to incorporate Snowflake Cortex AI Functions
into the target SQL. Constructing a benchmark around AI-native
SQL introduces two challenges that do not arise in traditional
text-to-SQL benchmarks. The first is \textbf{specification
determinism} at the instruction level: AI functions introduce
additional degrees of freedom beyond conventional SQL, including
which function to use and how to specify its parameters (e.g.,
the set of class labels for a classification task). Instructions
must be precise enough to determine these choices, or the notion
of a correct answer becomes ill-defined. The second is
\textbf{execution determinism} at the execution level: as discussed
in \S\ref{sec:aifunc}, AI functions invoke language models whose
outputs may vary slightly even under deterministic settings. The
gold SQL must therefore produce a stable result for
execution-based evaluation to be meaningful. Our construction
process (\S\ref{sec:construct}) addresses specification
determinism, and our verification protocol (\S\ref{sec:verify})
addresses execution determinism. We describe the task definition
in \S\ref{sec:task}, dataset statistics in \S\ref{sec:stats},
and evaluation setup in \S\ref{sec:eval}.

\subsection{Task Definition}
\label{sec:task}

The input to each Spider 2.0-AIFunc task consists of a natural
language instruction $Q$, a database schema $D$, auxiliary
documentation $E$, and AI function reference documentation $R$.
The first three components follow the same format as Spider2-Snow.
The additional component $R$ provides definitions, parameter
specifications, and usage guidelines for the relevant Snowflake
Cortex AI Functions, since these functions are relatively new and
unlikely to be well represented in the training data of current
language models. Given these inputs, a system $f(\cdot)$ (which
may be a text-to-SQL parser, an LLM, or an agent framework) is
required to produce a SQL query $S = f(Q, D, E, R \mid \theta)$
that includes one or more AI function calls alongside
conventional SQL operations. During construction, we refine
instructions to reduce ambiguity and make the intended AI
function usage and parameters explicit, as discussed in
\S\ref{sec:construct}.

\subsection{Task Construction}
\label{sec:construct}

Our construction starts from 513 Spider2-Snow source tasks.
For each task, we take the original instruction, one or more
gold SQL queries and their execution results, the database
schema, and optional external knowledge as input. The goal is
to produce an AI-native SQL version of the task by incorporating
one or more Snowflake Cortex AI Functions into the target SQL.
The six AI function types covered in our benchmark are:
\texttt{AI\_CLASSIFY} for semantic classification,
\texttt{AI\_FILTER} for filtering rows by semantic relevance,
\texttt{AI\_SENTIMENT} for sentiment analysis,
\texttt{AI\_SIMILARITY} for embedding-based similarity
computation, \texttt{AI\_EXTRACT} for extracting structured
information from text, and \texttt{AI\_AGG} for aggregating
content across rows using natural language instructions. These
represent a broad range of semantic operations that cannot be
expressed with conventional SQL operators. Note that each task
may incorporate more than one AI function, and most tasks in
our benchmark combine two functions within a single query.

The construction is carried out by a reusable, agent-based
pipeline using Claude Opus 4.5~\citep{anthropic_opus_45} with a
maximum of 15 interaction rounds per task. Agents interact directly with the databases, simultaneously proposing modifications to both the
SQL and the instruction. For the SQL, agents inject or rewrite
portions of the query to incorporate AI function calls. For the
instruction, agents revise the wording to match the rewritten
SQL; when a source task has multiple gold SQLs, the instruction
is further refined to reduce ambiguity and make the intended
query more explicit. Throughout this process, agents
execute the candidate SQL against the database and iteratively
resolve issues, including execution errors, timeouts, and
malformed outputs, until the query returns reasonable results
within an acceptable time frame. The agents also verify that
all AI function parameters are fully specified: for instance,
that a call to \texttt{AI\_CLASSIFY} includes the complete set
of target labels, or that a call to \texttt{AI\_EXTRACT}
specifies the exact extraction schema. This addresses
specification determinism at construction time by making the
intended AI-function choices and parameters explicit.

The benchmark is constructed in two rounds. In the main round,
agents select AI functions without constraints on function
choice. This produces 513 candidate instances, but the
resulting distribution is concentrated on three function types:
\texttt{AI\_CLASSIFY} appears in 78.6\% of tasks,
\texttt{AI\_FILTER} in 50.1\%, and \texttt{AI\_SIMILARITY} in
45.6\%, while \texttt{AI\_SENTIMENT} (0.2\%),
\texttt{AI\_EXTRACT} (1.2\%), and \texttt{AI\_AGG} (10.7\%)
remain rare. The two most frequent combinations,
\texttt{AI\_CLASSIFY} + \texttt{AI\_FILTER} (35.3\%) and
\texttt{AI\_CLASSIFY} + \texttt{AI\_SIMILARITY} (33.7\%),
together account for 69\% of all instances. This concentration
reflects the broad applicability of classification, filtering,
and similarity operations across diverse analytical tasks, but
it results in insufficient coverage of the remaining function
types. To address this, we conduct a diversity round in which
agents are explicitly instructed to prioritize
underrepresented functions: \texttt{AI\_SENTIMENT} as the
highest priority, \texttt{AI\_EXTRACT} as the second, and
\texttt{AI\_AGG} as the third. Agents are required to include
at least one of these target functions in each task, but may
reject a source instance if none of the target functions
naturally fits the data and query intent. This yields 227
additional candidate instances with a complementary
distribution: \texttt{AI\_AGG} appears in 71.4\% of
diversity-round tasks, \texttt{AI\_EXTRACT} in 31.3\%, and
\texttt{AI\_SENTIMENT} in 15.0\%. Instances from both rounds
then proceed to the verification stage (\S\ref{sec:verify});
only those that pass are retained in the final release. The
pipeline is designed to be general and can be adapted to
construct future benchmarks for other AI functions or database platforms.

\subsection{Determinism Verification}
\label{sec:verify}

After construction, each candidate instance undergoes a
multi-pass verification process to check execution
determinism. The process consists of four consecutive passes:
the first executes each gold SQL 5 times, and the subsequent
three each execute it 10 times, totaling at least 35 executions
per instance (additional executions may be triggered when
repairs require re-verification). All passes are carried out by
a verification agent (Claude Opus 4.5, maximum 15 interaction
rounds). The agent is permitted to repair instances that
produce inconsistent results: it may adjust the SQL, revise the
instruction, or modify evaluation configuration such as whether
row order should be compared. Beyond simple re-execution, the
agent can run diagnostic SQL queries against the database to
identify the source of inconsistencies before attempting a fix.
Importantly, the agent also reviews instances that already pass
consistency checks, examining whether the instruction
sufficiently determines the intended AI function usage and
whether the evaluation configuration is appropriate. This means
the verification stage checks execution determinism while also
reviewing residual specification-determinism concerns that may
have survived construction. An instance is considered consistent
only when 100\% of executions within a pass produce identical
results under the same comparison logic used for final
evaluation. After all four passes, we apply filtering: instances
that remain inconsistent are removed, as are instances whose AI
function usage was reduced or eliminated during repair.
We additionally discard instances built on a small number of
Spider2-Snow source databases that are not static snapshots and whose
underlying tables may be updated over time, since such changes can
alter the inputs passed to AI functions and make results unstable
across evaluation periods.

The surviving instances then enter a final stability check.
Each instance is executed 10 times in each of three separate
time windows, for a total of 30 additional executions. No repairs are made during this stage; any instance whose 10
executions within any single time window produce inconsistent
results is dropped. This
stage serves a different purpose from the repair-based passes:
it checks that the released instances remain stable not only
within a single session but also across temporally
separated evaluations. After all verification and filtering
stages, 465 instances (325 from the main round, 140 from the
diversity round) across 125 databases constitute the final
released benchmark.

\subsection{Dataset Statistics}
\label{sec:stats}

\begin{figure}[t]
\centering
\includegraphics[width=0.96\textwidth]{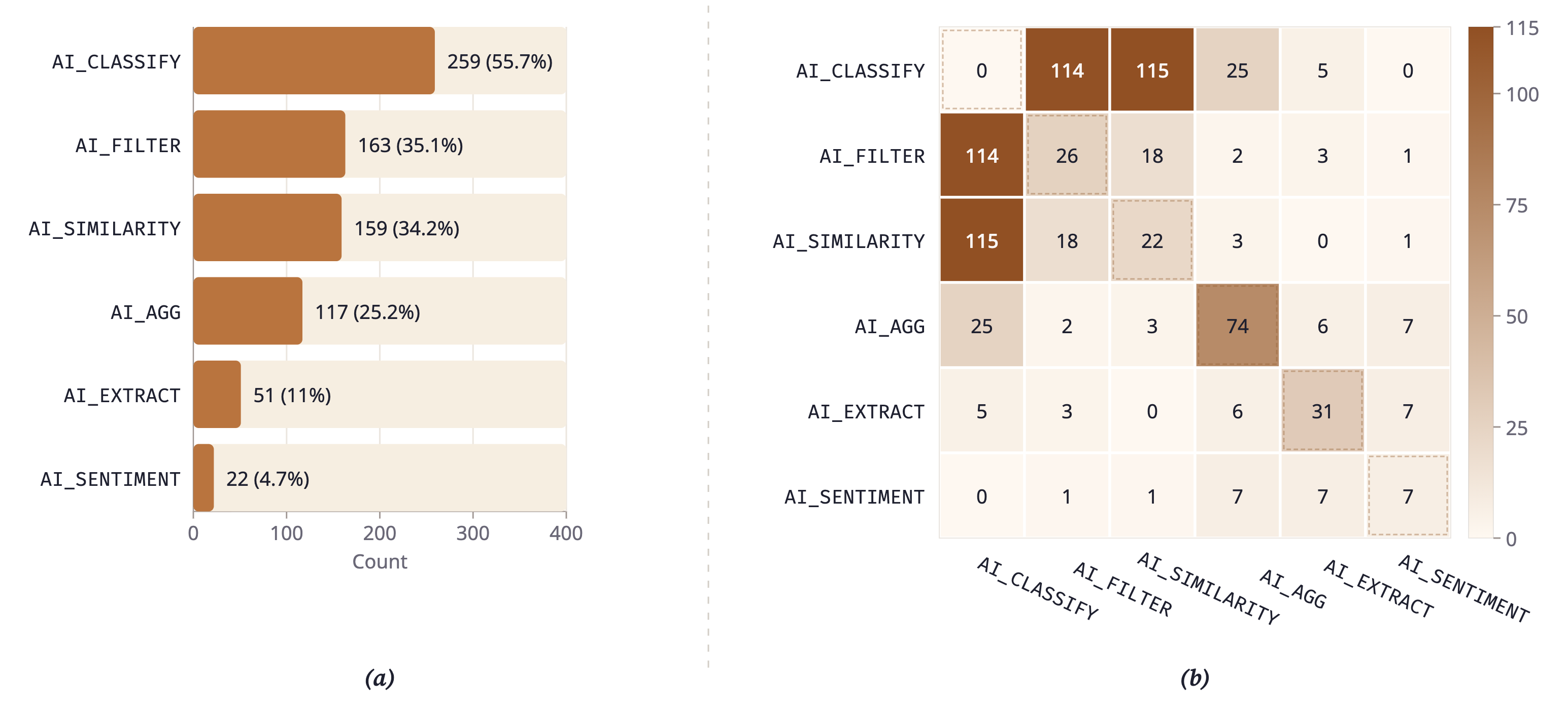}
\caption{(a) Distribution of AI function types across the
465 released instances. A single task may use multiple
functions. (b) Function co-occurrence matrix. Diagonal
entries indicate single-function tasks; off-diagonal entries
indicate two-function combinations.}
\label{fig:func_dist}
\vspace{-0.4cm}
\end{figure}

The final benchmark comprises 465 verified instances across
125 databases. Figure~\ref{fig:func_dist}(a) shows the
distribution of AI function types across all released
instances. \texttt{AI\_CLASSIFY} is the most frequently used
function, appearing in over half of all tasks, followed by
\texttt{AI\_FILTER} and \texttt{AI\_SIMILARITY} at comparable
rates. The diversity round substantially increased the
representation of \texttt{AI\_AGG}, \texttt{AI\_EXTRACT}, and
\texttt{AI\_SENTIMENT}, which together now appear in over 40\%
of tasks. Two thirds of the tasks (66.0\%) combine two AI
functions within a single query, while 33.3\% use a single
function. Figure~\ref{fig:func_dist}(b) presents the
co-occurrence matrix across the six function types.
Since Spider 2.0-AIFunc is built from Spider2-Snow source
tasks, the gold SQL queries retain the characteristics of
enterprise-level SQL: multi-table joins, common table
expressions, and layered aggregations. AI function calls are
embedded within these multi-step queries rather than appearing
in isolation, meaning that a system must not only select and
parameterize the correct AI functions but also integrate them
into the surrounding query logic to produce the intended
result. The gold queries average 65 lines and 2,617 characters
in length, referencing an average of 3.9 tables per query. The
gold query results average 23.1 rows (median 3) and 3.7
columns (median 3), with 91.8\% of tasks returning 20 rows or
fewer and 2.2\% exceeding 100 rows. Execution times average
10.4 seconds (median 4.3 seconds), with 2.2\% of instances
requiring over a minute due to AI function invocations across
large numbers of rows.

\subsection{Evaluation}
\label{sec:eval}

We evaluate using execution accuracy, following the framework
of Spider2-Snow. In Spider2-Snow, the gold SQL
result is pre-computed and stored; at evaluation time, only the
predicted SQL is executed and compared against the stored
result. In Spider 2.0-AIFunc, this approach is not viable:
AI functions invoke language models that may be updated over
time (\S\ref{sec:aifunc}), so a stored gold result may no
longer reflect the current output of the same gold SQL. We
therefore require that the gold SQL $S_n$ and the predicted SQL
$\hat{S}_n$ are both executed within the same time window $t$:
\begin{equation}
EX = \frac{1}{N}\sum_{n=1}^{N}
      \mathbb{1}\!\Big(
        \mathrm{exec}(S_n,\, t),\;
        \mathrm{exec}(\hat{S}_n,\, t)
      \Big)
\end{equation}
where $\mathrm{exec}(S, t)$ denotes the result of executing
query $S$ at time $t$, and the shared $t$ ensures that both
queries are evaluated against the same underlying model state.
Let $v = \mathrm{exec}(S_n, t)$ and
$\hat{v} = \mathrm{exec}(\hat{S}_n, t)$. Following
Spider2-Snow, the comparison function $\mathbb{1}(v, \hat{v})$
uses column-subset matching:
\begin{equation}
\mathbb{1}(v, \hat{v}) =
  \begin{cases}
    1 & \text{if } v_i \in \hat{v},\;
        \forall\, v_i \in v \\
    0 & \text{if } v_i \notin \hat{v},\;
        \exists\, v_i \in v
  \end{cases}
\end{equation}
where $v_i$ denotes the $i$-th column of result $v$. That is,
every column in the gold result must appear in the predicted
result, but the predicted result may contain additional columns
without penalty. Each instance additionally carries an
evaluation configuration that specifies whether row order
should be compared or ignored. Float values are compared with
a tolerance of 0.01, and columns are matched by content rather
than by position. This comparison logic is identical to the one
used during the verification stage (\S\ref{sec:verify}).

\section{Experiments}

\subsection{Experimental Setup}
\label{sec:setup}

We evaluate all models using the Spider-Agent framework~\citep{lei2025spider20evaluatinglanguage}, with a minimal tool set consisting
of three general-purpose tools: a bash tool for exploring
database schemas and external knowledge files, a SQL execution
tool for running queries directly against the Snowflake
environment, and a task termination tool for signaling
completion. We deliberately keep the tool set minimal to isolate
model capability from framework complexity. Compared to the
original Spider-Agent setup, the primary addition on the task
side is the inclusion of AI function reference documentation
$R$ (defined in \S\ref{sec:task}), which provides each model
with definitions, parameter specifications, and usage guidelines
for the relevant Snowflake Cortex AI Functions.
The full prompt,
including the reference documentation and usage guidelines, is
provided in Appendix~\ref{app:prompt}.
All agents are run with a maximum of 50 interaction rounds,
a temperature of 1, and a per-query execution timeout of 120
seconds. For each model and task, we report one complete agent
trajectory. To account for transient warehouse instability that
may cause spurious timeouts, each predicted SQL is executed
up to three times; a query is marked as timed out only if
all three attempts exceed the timeout threshold. For models that expose a reasoning effort parameter,
we set it to \texttt{medium} uniformly across all such models;
models that do not support this parameter are run with their
default configuration.

\begin{table}[t]
\centering
\setlength{\tabcolsep}{5pt}
\renewcommand{\arraystretch}{1.35}
\caption{%
  Execution accuracy on \textsc{Spider~2.0-AIFunc}.
  \textit{Exec.} reports the proportion of instances on which the
  predicted SQL ran without error or timeout. \textit{EX} is execution accuracy.
  Best results within each group are highlighted. $^\dagger$GPT-5.4 results are reported with model-specific
  patches applied to address specific issues
  observed during evaluation; see Appendix~\ref{app:gpt} for
  details.
}
\label{tab:main}
\begin{tabular}{@{} l r r r l @{}}
\toprule
\textbf{Model} &
\textbf{Params} &
\textbf{Exec.~(\%)} &
\textbf{EX~(\%)} &
\\
\midrule

\multicolumn{5}{@{}l}{\grouphdr{Proprietary Models}{grouporange}}\\[3pt]

\mlogo{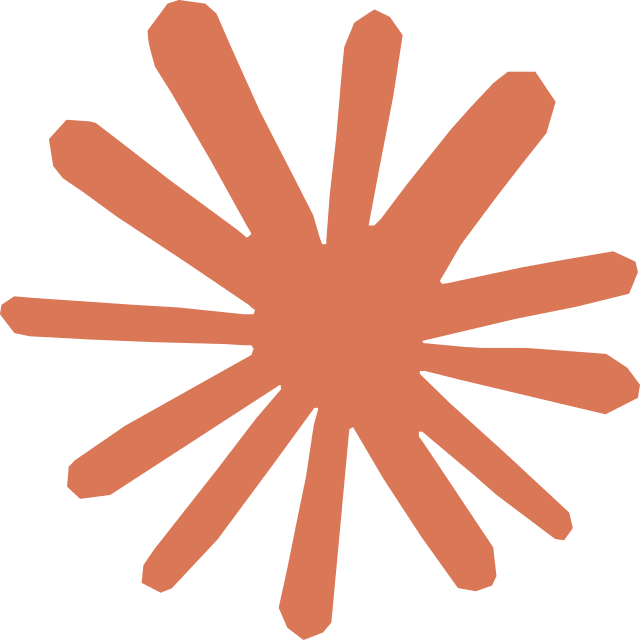}~Claude Opus 4.6   &
-- &
99.4 & \bestclosed{70.3} &
\pbarrounded{70.3}{closedorange}{closedorangelight} \\

\mlogo{anthropic.png}~Claude Sonnet 4.6 &
-- &
98.9 & 69.0 &
\pbarrounded{69.0}{closedorange}{closedorangelight} \\

\mlogo{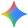}~Gemini 3.1 Pro       &
-- &
97.8 & 67.1 &
\pbarrounded{67.1}{closedorange}{closedorangelight} \\

\mlogo{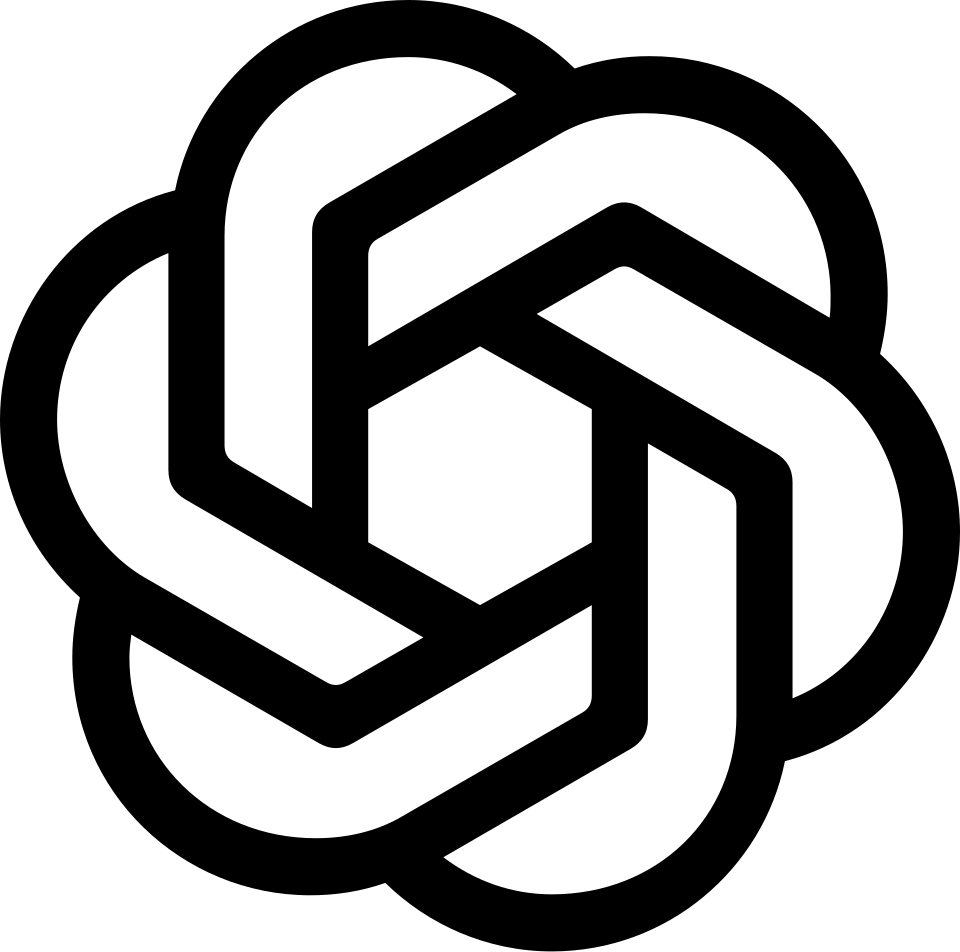}~GPT-5.4$^\dagger$              &
-- &
97.6 & 63.0 &
\pbarrounded{63.0}{closedorange}{closedorangelight} \\

\mlogo{google.png}~Gemini 3 Flash       &
-- &
90.6 & 60.9 &
\pbarrounded{60.9}{closedorange}{closedorangelight} \\

\midrule
\multicolumn{5}{@{}l}{\grouphdr{Open-source Models}{groupteal}}\\[3pt]

\mlogo{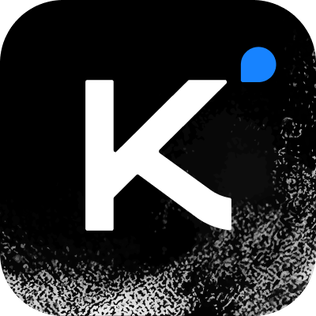}~Kimi K2.5              &
1000B / 32B &
97.8 & \bestopen{58.1} &
\pbarrounded{58.1}{openteal}{openteallight} \\

\mlogo{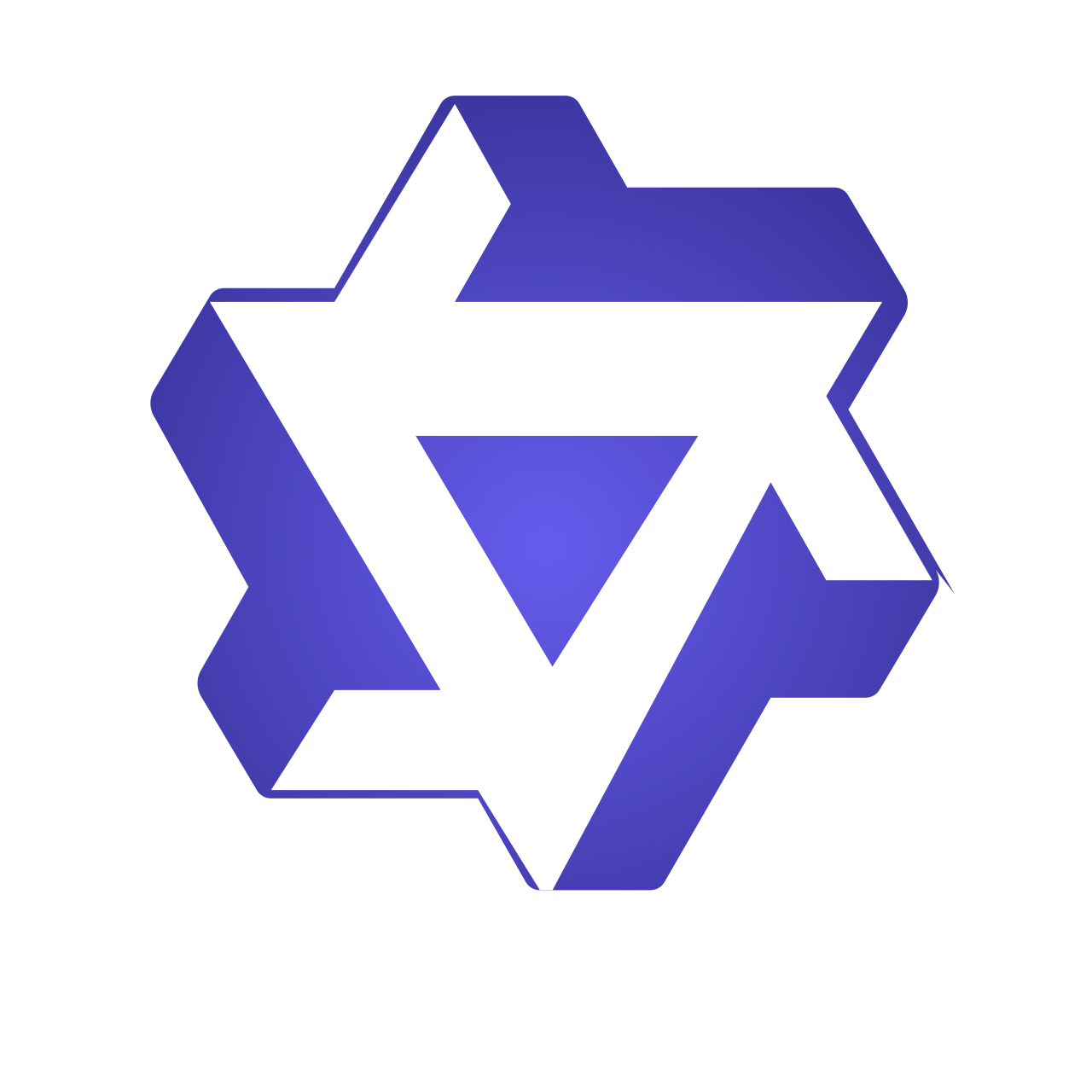}~Qwen 3.5 Plus          &
397B / 17B &
97.2 & 57.0 &
\pbarrounded{57.0}{openteal}{openteallight} \\

\mlogo{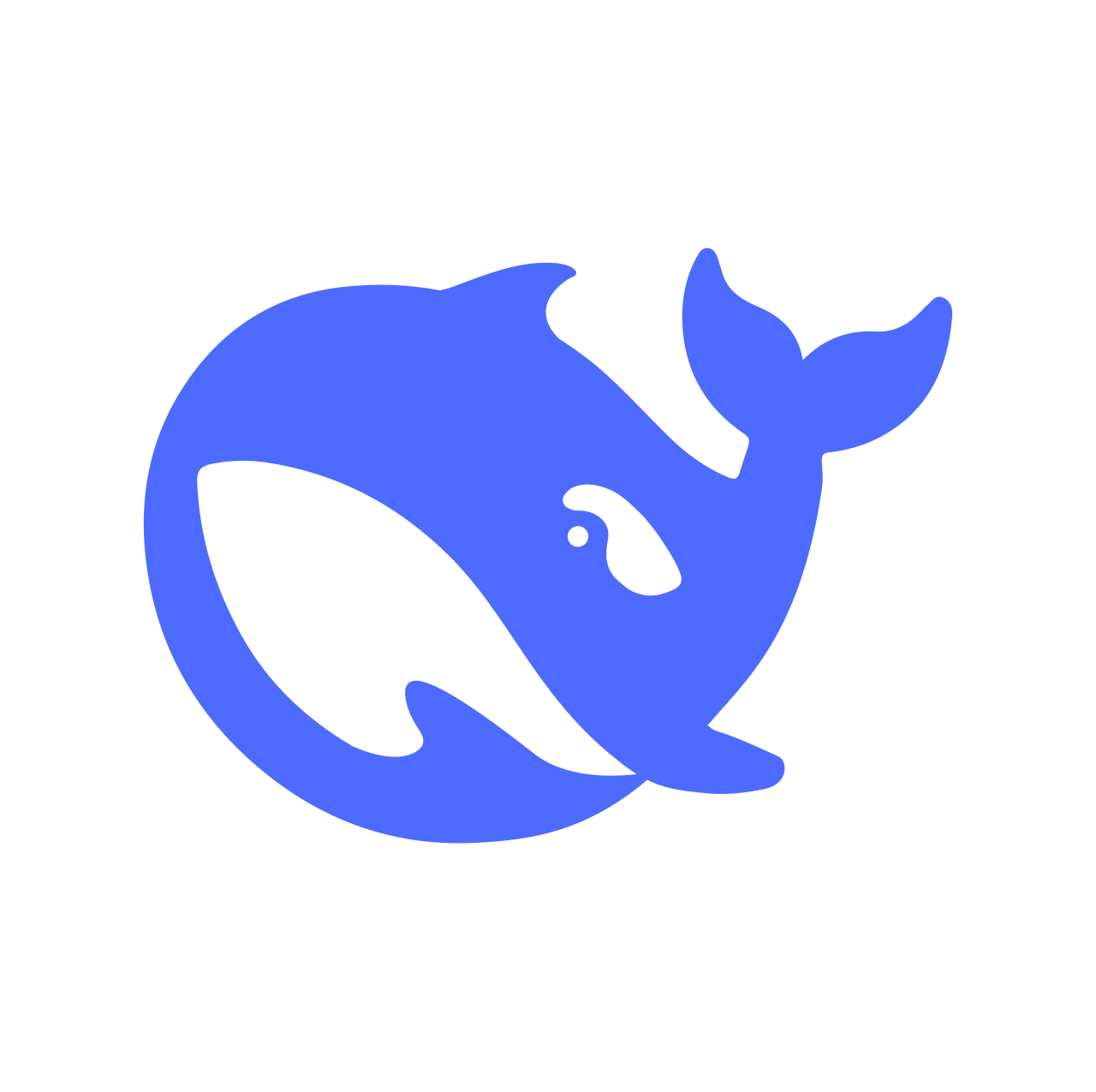}~DeepSeek V3.2      &
671B / 37B &
97.2 & 54.6 &
\pbarrounded{54.6}{openteal}{openteallight} \\

\mlogo{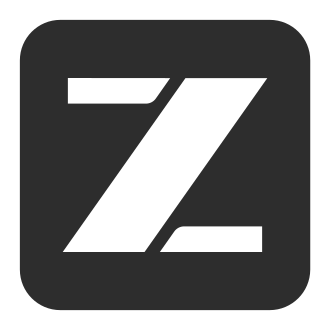}~GLM-5                 &
744B / 40B &
94.6 & 52.9 &
\pbarrounded{52.9}{openteal}{openteallight} \\

\mlogo{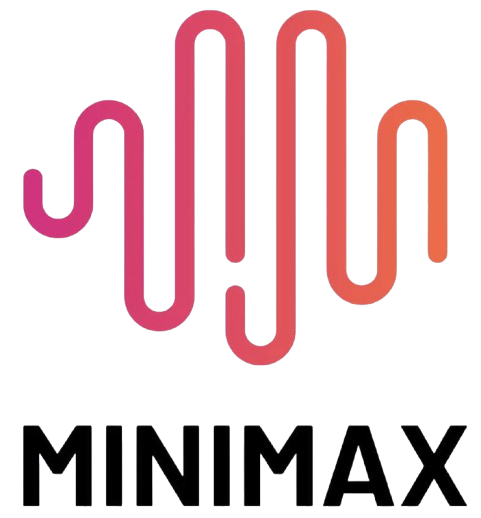}~MiniMax M2.5        &
230B / 10B &
96.1 & 44.9 &
\pbarrounded{44.9}{openteal}{openteallight} \\

\bottomrule
\end{tabular}
\end{table}

We evaluate ten models spanning both closed-source and
open-source families. On the closed-source side, we include
Claude Opus 4.6~\citep{anthropic_opus_46} and Claude Sonnet 4.6~\citep{anthropic_sonnet_46}, Gemini 3.1
Pro~\citep{google_gemini_31_pro} and Gemini 3 Flash~\citep{google_gemini_3_flash}, and GPT-5.4~\citep{openai_gpt54}. On the
open-source side, we include Kimi K2.5~\citep{kimiteam2026kimik25visualagentic}, Qwen 3.5 Plus~\citep{qwen35_plus},
DeepSeek V3.2~\citep{deepseekai2025deepseekv32pushingfrontieropen}, GLM-5~\citep{glm5team2026glm5vibecodingagentic}, and MiniMax M2.5~\citep{minimax_m25}. We additionally compare three open-sourced agent frameworks,
AutoLink~\citep{wang2025autolinkautonomousschemaexploration}, ReFoRCE~\citep{deng2025reforcetexttosqlagentselfrefinement}, and DSR-SQL~\citep{hao2025texttosqldualstatereasoningintegrating}, each using Claude Sonnet 4.6
as the backbone model under their default configurations, to
assess whether framework complexity confers any advantage on
this benchmark. To ensure a fair comparison, we augment each
framework with the same AI function reference documentation
$R$ used in the Spider-Agent setting, inserting it into all
available system prompts across each framework's internal
stages; all other configurations are left unchanged.

\subsection{Main Results}
\label{sec:results}

\textbf{Models generally produce executable AI-native SQL
across sufficient interaction rounds.} Table~\ref{tab:main}
reports the results across all ten models. We measure two
metrics: \textit{Exec.}, the proportion of instances on which
the predicted SQL ran without error, and execution accuracy
(\textit{EX}), which additionally requires the query to return
the correct result. Across all models, \textit{Exec.} is
consistently high, ranging from 90.6\% to 99.4\%, suggesting
that with sufficient interaction rounds, models can generally
produce syntactically valid AI-native SQL that executes without
error or timeout.

\textbf{Proprietary models consistently outperform open-source
models.} Among proprietary models, Claude Opus 4.6 achieves
the highest accuracy at 70.3\%, followed by Claude Sonnet 4.6
at 69.0\% and Gemini 3.1 Pro at 67.1\%. A notable gap emerges
between proprietary and open-source models: the strongest
proprietary models cluster in the 67--70\% range, while the
best open-source model, Kimi K2.5, reaches 58.1\% and the
remaining open-source models range from 44.9\% to 57.0\%.

\textbf{Agent frameworks designed for traditional text-to-SQL
do not transfer effectively to AI-native SQL workflows.}
Table~\ref{tab:framework} compares Spider-Agent against three
alternative agent frameworks, all using Claude Sonnet 4.6 as
the backbone model under their default configurations and
ranked by \textit{EX}. Spider-Agent achieves 69.0\%, matching
AutoLink (68.6\%) and ReFoRCE (68.5\%), while DSR-SQL trails
at 62.2\% despite achieving the highest \textit{Exec.} of
100\%. These frameworks
were originally designed for traditional text-to-SQL challenges
such as schema retrieval and relevant table selection; their
relative ineffectiveness here may reflect that such strategies
are less critical as model capabilities improve, or that they
do not transfer naturally to AI-native SQL workflows.

Our qualitative inspection of disagreement cases suggests that these
frameworks do transfer partially, but unevenly. They can help with
conventional text-to-SQL components such as schema context, subtle
filters, or multi-step SQL templates. However, AI-native SQL correctness
often depends on small semantic decisions introduced by the AI-function
call itself: which rows are passed into the function, whether filtering
happens before or after the AI call, the aggregation grain,
semi-structured field paths, and the exact function parameters.
Decomposition, context selection, or intermediate rewriting can therefore
introduce small deviations that offset the benefits of stronger
traditional SQL scaffolding.

Detailed per-function-type breakdowns, single- versus multi-function accuracy,
AI function selection accuracy, and execution time analysis are reported in
Appendix~\ref{app:analysis}.

\begin{table}[t]
\centering
\setlength{\tabcolsep}{8pt}
\renewcommand{\arraystretch}{1.35}
\caption{%
  Agent framework comparison on \textsc{Spider~2.0-AIFunc}.
  All frameworks use Claude Sonnet 4.6 as the backbone model.
}
\label{tab:framework}
\begin{tabular}{@{} l r r @{}}
\toprule
\textbf{Framework} &
\textbf{Exec.~(\%)} &
\textbf{EX~(\%)} \\
\midrule
Spider-Agent   & 98.9 & 69.0 \\
AutoLink~\citep{wang2025autolinkautonomousschemaexploration}      & 98.7 & 68.6          \\
ReFoRCE~\citep{deng2025reforcetexttosqlagentselfrefinement}       & 99.8 & 68.5          \\
DSR-SQL~\citep{hao2025texttosqldualstatereasoningintegrating}       & 100.0 & 62.2         \\
\bottomrule
\end{tabular}
\end{table}

\subsection{Agent Interaction Behavior}
\label{sec:interaction}

Table~\ref{tab:interaction} reports how each model distributes its
50-round budget across the two primary tools in our Spider-Agent
setup: \texttt{execute\_sql} for running queries against Snowflake,
and \texttt{bash} for exploring database schemas and external
knowledge files. \textit{1st SQL} and \textit{1st Succ.}\ denote the
average round at which the model first calls \texttt{execute\_sql}
and first produces an error-free execution, respectively.

Bash call counts are stable across models, ranging from 4.1 to 7.8,
while SQL execution counts span a substantially wider range, from 4.4
for Gemini~3~Flash to 19.8 for DeepSeek~V3.2. The gap in total rounds
between proprietary and open-source models is therefore driven almost
entirely by how many times each model executes SQL rather than how
much it explores. Proprietary models average 4--6 SQL calls per
instance and typically exhaust their budget at around 10 rounds;
open-source models average 12--20 SQL calls over 19--30 rounds, with
comparable bash usage. Open-source models also defer their first SQL
attempt to around rounds 7--8, compared to rounds 5--6 for proprietary
models, but this 2-round difference accounts for only a small fraction
of the overall gap. The bulk of the difference accumulates after the
first successful execution, suggesting that open-source models require
substantially more iterative refinement to arrive at a correct result.

Open-source models are also more likely to exhaust the 50-round
budget without terminating: MiniMax~M2.5 reaches the limit on 10.8\%
of instances and GLM-5 on 6.2\%, compared to at most 0.6\% for any
proprietary model. Beyond their tendency to consume more rounds,
we observe that a fraction of these cases stem from malformed
\texttt{terminate} tool calls, where the model repeatedly issues
incorrectly formatted termination tool calls without recognizing the error,
preventing the agent from signaling completion.

\begin{table}[t]
\centering
\setlength{\tabcolsep}{5.5pt}
\renewcommand{\arraystretch}{1.35}
\caption{%
  Agent interaction statistics averaged over all 465 instances.
  \textit{Rounds}: average total interaction rounds.
  \textit{SQL} and \textit{Bash}: average number of
  \texttt{execute\_sql} and \texttt{bash} calls.
  \textit{B/S}: bash-to-SQL call ratio.
  \textit{Rnd-50}: proportion of instances reaching the
  50-round limit.
  \textit{1st SQL}: average round of the first
  \texttt{execute\_sql} call.
  \textit{1st Succ.}: average round of the first
  error-free SQL execution.
}
\label{tab:interaction}
\begin{tabular}{@{} l @{\hskip 10pt} r r r r r @{\hskip 10pt} r r @{}}
\toprule
& \multicolumn{5}{c}{\textit{Interaction volume (avg.\ per instance)}}
& \multicolumn{2}{c}{\textit{Timing (avg.\ round)}} \\
\cmidrule(lr){2-6}\cmidrule(l){7-8}
\textbf{Model}
& \textbf{Rounds}
& \textbf{SQL}
& \textbf{Bash}
& \textbf{B/S}
& \textbf{Rnd-50 (\%)}
& \textbf{1st SQL}
& \textbf{1st Succ.} \\
\midrule

\multicolumn{8}{@{}l}{\grouphdr{Proprietary Models}{grouporange}}\\[3pt]

\mlogo{anthropic.png}~Claude Opus 4.6
  & 11.7 & 5.9  & 4.9 & 1.09 &  0.0 & 5.7 & 5.7 \\
\mlogo{anthropic.png}~Claude Sonnet 4.6
  & 10.0 & 4.9  & 4.1 & 1.42 &  0.0 & 4.9 & 5.3 \\
\mlogo{google.png}~Gemini 3.1 Pro
  & 17.1 & 10.2 & 5.8 & 0.79 &  0.6 & 5.6 & 6.6 \\
\mlogo{openai.png}~GPT-5.4$^\dagger$
  & 10.1 & 4.4  & 4.4 & 1.45 &  0.0 & 5.1 & 5.6 \\
\mlogo{google.png}~Gemini 3 Flash
  &  9.9 & 4.4  & 4.3 & 1.47 &  0.0 & 4.7 & 5.1 \\

\midrule
\multicolumn{8}{@{}l}{\grouphdr{Open-source Models}{groupteal}}\\[3pt]

\mlogo{kimi.png}~Kimi K2.5
  & 19.3 & 12.1 & 6.1 & 0.74 &  0.4 & 6.9 & 7.3 \\
\mlogo{qwen.png}~Qwen 3.5 Plus
  & 24.1 & 16.4 & 6.8 & 0.54 &  3.2 & 7.4 & 8.0 \\
\mlogo{deepseek.png}~DeepSeek V3.2
  & 28.6 & 19.8 & 7.8 & 0.45 &  0.4 & 7.8 & 8.2 \\
\mlogo{zhipu.png}~GLM-5
  & 24.5 & 14.4 & 7.0 & 0.94 &  6.2 & 8.0 & 8.4 \\
\mlogo{minimax.png}~MiniMax M2.5
  & 30.0 & 19.1 & 6.7 & 0.47 & 10.8 & 7.7 & 8.7 \\

\bottomrule
\end{tabular}
\end{table}

\subsection{Error Analysis}
\label{sec:error}

To better understand model failures on \textsc{Spider
2.0-AIFunc}, we conduct a stratified error analysis over a
random sample of 85 instances drawn from three strata defined
by which models succeed. \textbf{S1} ($n=30$) contains instances where all three strong proprietary
models (Claude Opus 4.6, Gemini 3.1 Pro, and GPT-5.4) fail.
\textbf{S2} ($n=30$) contains
instances where at least one strong proprietary model succeeds.
\textbf{S3} ($n=25$) contains
instances where all three strong proprietary models succeed
but at least one open-source model fails. For each sampled
instance, we manually identify the primary error type(s)
from the predicted SQL of failing models, categorized into
five types: \textbf{C1} (schema grounding: wrong table,
column, or prefix selection), \textbf{C2} (predicate \&
filter: missing or incorrect filtering conditions),
\textbf{C3} (query logic: incorrect aggregation granularity,
join scope, or algorithmic errors), \textbf{C4} (AI function
usage: wrong function, incorrect syntax, missing or
misspecified parameters), and \textbf{C5} (annotation issues:
ambiguous or incorrect gold queries). Instances may carry
multiple error labels; full category definitions and
representative examples are provided in
Appendix~\ref{app:error}. Figure~\ref{fig:error} summarizes the error distribution
across the three strata.

\begin{figure}[t]
\centering
\includegraphics[width=\textwidth]{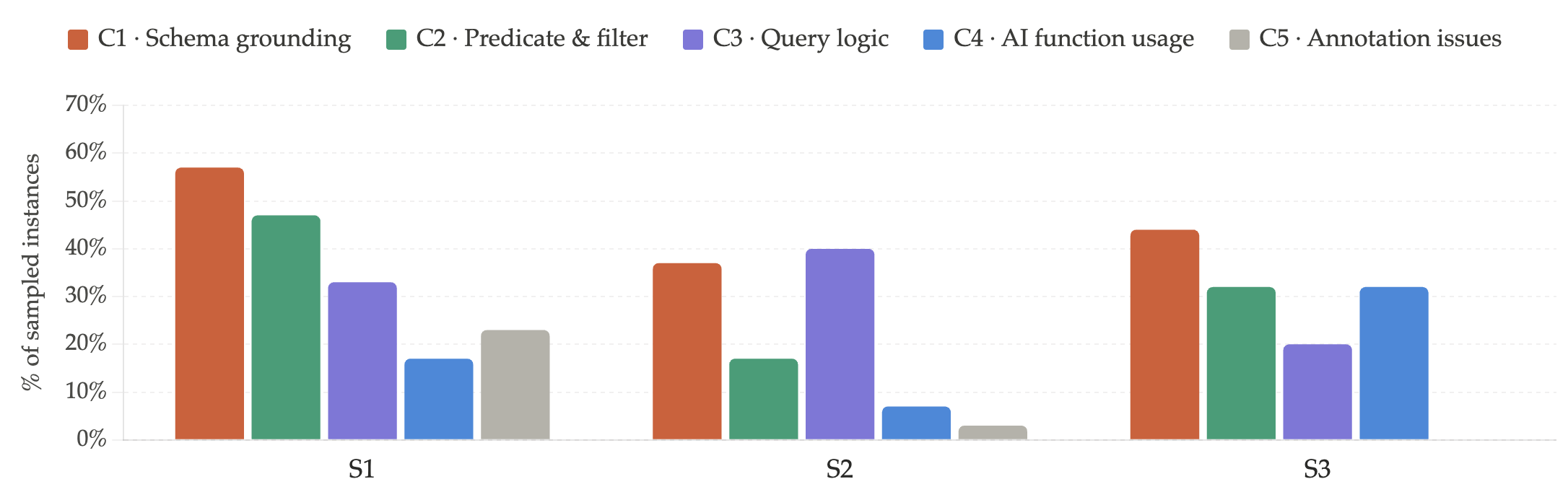}
\caption{%
  Distribution of error types across three failure strata,
  based on manual analysis of randomly sampled instances.
  \textbf{S1}: all three strong proprietary models fail;
  \textbf{S2}: at least one strong proprietary model succeeds;
  \textbf{S3}: all strong proprietary models succeed but at
  least one open-source model fails.
  Error categories: \textbf{C1} schema grounding,
  \textbf{C2} predicate \& filter,
  \textbf{C3} query logic,
  \textbf{C4} AI function usage,
  \textbf{C5} annotation issues (gold query ambiguity or
  errors; reported for S1 only).
  Instances may carry multiple error labels.
}
\vspace{-0.3cm}
\label{fig:error}

\end{figure}

\textbf{Universal failures are driven by predicate and
schema errors.} In S1, where all strong models fail, C2
(predicate \& filter) and C1 (schema grounding) are the most
prevalent error types, appearing in 47\% and 57\% of sampled
instances respectively, often co-occurring. These errors
reflect cases where the instruction does not fully specify
filtering conditions or where the correct table and column
names require precise knowledge of the database schema that
all models lack. Notably, 7 of the 30 S1 instances (23\%)
show signs of annotation issues (C5), where the gold query
encodes constraints absent from the instruction or admits
multiple valid answers, suggesting that a non-trivial portion
of universal failures may reflect benchmark limitations
rather than model deficiencies.

\textbf{Aggregation and query logic errors distinguish strong
models from weaker ones.} In S2, where at least one strong
model succeeds, C3 (query logic) is the dominant error type,
appearing in 40\% of instances. Failures in this stratum tend
to involve subtle decisions such as when to apply filtering
before or after an AI function call, how to handle nested
aggregations, or how to correctly process semi-structured
data types. These are errors that stronger models more
reliably avoid, suggesting that C3 captures a meaningful
dimension of model capability on AI-native SQL tasks.

\textbf{Open-source models show systematic weaknesses in
schema grounding and AI function usage.} In S3, where all
strong proprietary models succeed but open-source models
fail, C1 (schema grounding) and C4 (AI function usage) each
appear in roughly 32\% of instances, and C4 is nearly absent
from S1 and S2. This indicates that non-standard AI function
syntax and incorrect parameter specifications are
disproportionately an open-source model failure mode,
consistent with the function selection accuracy results
in Appendix~\ref{app:func_select}. Schema grounding errors
in S3 tend to involve subtler mistakes than in S1, such as
selecting a table from the wrong schema family or
misidentifying column aliases, which stronger models resolve
through more thorough exploration of the database environment.

\section{Related Work}

The landscape of text-to-SQL evaluation has evolved from early datasets targeting domain generalization and lexico-logical alignments, such as KaggleDBQA~\citep{lee2021kaggledbqarealisticevaluationtexttosql} and SQUALL~\citep{shi2020potentiallexicologicalalignmentssemantic}, to highly complex, enterprise-grade environments. Recent benchmarks like BEAVER~\citep{chen2025beaverenterprisebenchmarktexttosql} and Spider 2.0~\citep{lei2025spider20evaluatinglanguage} introduce extreme schema complexity and multi-step real-world workflows, while datasets like MultiSpider 2.0~\citep{pham2025multilingualtexttosqlbenchmarkinglimits} and CORGI~\citep{li2026agentbainvsagent} extend these challenges to multilingual settings and predictive business logic. To navigate these increasingly massive schemas, researchers have developed sophisticated multi-agent reasoning frameworks, such as ReFoRCE~\citep{deng2025reforcetexttosqlagentselfrefinement}, AutoLink~\citep{wang2025autolinkautonomousschemaexploration}, DSR-SQL~\citep{hao2025texttosqldualstatereasoningintegrating}, and APEX-SQL~\citep{cao2026apexsqltalkingdataagentic}, that use iterative schema linking and agentic exploration. However, these evaluations and frameworks remain strictly confined to traditional relational algebra, overlooking a major shift in the database industry: the native integration of generative AI capabilities into query execution engines. Systems like LOTUS~\citep{patel2025semanticoperatorsdeclarativemodel} and MOAR~\citep{wei2026multiobjectiveagenticrewritesunstructured} have introduced semantic operators as declarative primitives, and commercial platforms such as BigQuery~\citep{bigquery_ai_intro}, Databricks~\citep{databricks_ai_functions}, and Snowflake Cortex~\citep{snowflake_cortex_ai_functions} now allow users to execute large language model operations directly within SQL statements. Spider 2.0-AIFunc addresses this gap by providing the first benchmark specifically tailored to evaluate how effectively modern models and agents can generate these emerging AI-native SQL workflows.

\section{Conclusion}

We introduced Spider 2.0-AIFunc, a benchmark for evaluating text-to-SQL
systems on AI-native SQL, in which target queries incorporate
Snowflake Cortex AI functions alongside conventional relational
operators. The benchmark comprises 465 verified instances across 125
real-world databases, constructed through an agent-based pipeline that
rewrites existing enterprise text-to-SQL tasks into AI-native form and
subjects every instance to a multi-round execution verification protocol
before release. Evaluation of ten state-of-the-art language models
shows a consistent capability gap between proprietary and open-source
systems, with errors concentrated in schema grounding, predicate
specification, query logic, and AI function parameterization.

\section{Limitations and Future Work}

Spider 2.0-AIFunc has several limitations. First, it is scoped to
Snowflake and the six Cortex AI function types available at the time of
construction, so it does not cover analogous functions on BigQuery,
Databricks, or other platforms. Second, although our construction
pipeline explicitly refines instructions to reduce ambiguity and make
the intended AI-function usage clear, it is difficult to fully verify
before release that every instance rules out all reasonable alternative
interpretations. As in Spider 2.0, future submissions may reveal
additional valid solutions or residual underspecification not captured
by the current reference SQL and evaluation configuration; we view these
submissions and issue reports as an important stress test for further
refining the benchmark. Third, our construction pipeline uses a single
strong construction model. Although each instance is derived from an
existing Spider2-Snow task with a real database, schema, and reference
SQL, this design may still introduce biases in wording style, rewrite
patterns, or AI-function choices.

Our evaluation is also limited by cost. We report a single agent
trajectory per model and task, while repeated sampling with pass@k or
variance estimates would provide a fuller picture of stochastic agent
performance. Similarly, repeated gold-prediction execution with majority
or stability-aware comparison could better capture AI-function output
variability. Both directions would substantially increase
model-inference cost and Snowflake AI-function execution cost, making
them impractical for the current study. Looking ahead, future work
should extend the construction pipeline to other platforms and function
families, incorporate stronger human or multi-generator validation, and
construct harder benchmarks that probe longer reasoning chains and more
diverse AI-function compositions.

\bibliography{colm2026_conference}
\bibliographystyle{colm2026_conference}

\newpage
\appendix
\section*{Appendix}

\section{Model-Specific Evaluation Adjustments for GPT-5.4}
\label{app:gpt}

GPT-5.4 exhibited two model-specific output patterns during
evaluation, which is why its result in Table~\ref{tab:main}
is marked with $^\dagger$ and accompanied by the ablation below.

\textbf{HTML entity encoding.}
Because our agent framework uses XML tags for tool calls,
GPT-5.4 frequently encoded SQL operators as HTML entities
(e.g., \texttt{\&gt;}, \texttt{\&lt;}, \texttt{\&amp;})
rather than their literal characters. This occurred in 184
of 465 instances, producing a total of 672 such entities,
and caused direct SQL compilation failures. No other model
exhibited this behavior. We applied automatic unescaping of
HTML entities prior to SQL execution uniformly across all runs reported in this paper.

\textbf{Parameter optimization.}
Despite instructions explicitly stating to strictly follow AI function
parameters as specified, GPT-5.4 consistently attempted to
rephrase or optimize prompt text, category labels, and
condition strings in AI function calls. Since evaluation
requires exact semantic alignment with the gold query, such
modifications cause evaluation failure. To address this, we
applied an additional patch to the system prompt:
\begin{quote}
\small\ttfamily
IMPORTANT: You MUST use the AI function parameters (category
labels, prompt text, condition strings, etc.) EXACTLY as
specified in the instruction. Any attempt to ``improve'',
rephrase, expand, or optimize these parameters will cause
evaluation failure.
\end{quote}
Table~\ref{tab:gpt_ablation} reports execution accuracy
across all combinations of patch status and reasoning effort
setting. The patch consistently improves accuracy by
approximately 10 percentage points regardless of reasoning
effort level.

We additionally conducted an ablation study
over reasoning effort settings (\texttt{medium},
\texttt{high}, \texttt{xhigh}). Contrary to expectations,
increasing reasoning effort yields no consistent improvement
in either the patched or unpatched setting, suggesting that
higher reasoning effort does not benefit AI-native SQL
generation for GPT-5.4 on this benchmark. We report the patched
\texttt{medium} result (63.0\%) in Table~\ref{tab:main}, and
include the full ablation here to make the effect of the adjustment
explicit.

\newcommand{\deltapill}[1]{%
  \colorbox{orangebg}{%
    \textcolor{closedorange}{%
      \small\bfseries\strut\;#1\;%
    }%
  }%
}

\begin{table}[htbp]
\centering
\setlength{\tabcolsep}{8pt}
\renewcommand{\arraystretch}{1.4}
\caption{%
  GPT-5.4 execution accuracy under different reasoning effort
  settings, with and without the parameter optimization patch.
  HTML entity unescaping is applied uniformly across all runs.
  $\Delta$ denotes the improvement from applying the patch.
  The result reported in Table~\ref{tab:main} is
  \textbf{bolded}.
}
\label{tab:gpt_ablation}
\begin{tabular}{@{} l rr c @{}}
\toprule
& \multicolumn{2}{c}{\textbf{EX (\%)}} & \\
\cmidrule(lr){2-3}
\textbf{Reasoning effort} &
\textbf{w/o patch} &
\textbf{w/ patch} &
\textbf{$\Delta$ (pp)} \\
\midrule
\texttt{medium} & 52.9 & \textbf{63.0} & \deltapill{$+10.1$} \\
\texttt{high}   & 51.2 & 62.2          & \deltapill{$+11.0$} \\
\texttt{xhigh}  & 51.4 & 61.9          & \deltapill{$+10.5$} \\
\bottomrule
\end{tabular}
\end{table}

\section{Detailed Analysis}
\label{app:analysis}

\subsection{AI Function Selection Accuracy}
\label{app:func_select}

Beyond execution accuracy, we measure whether models select
the correct set of AI functions for each task, defined as an
exact match between the predicted and gold function sets:
\begin{equation}
\mathrm{FSA} = \frac{|\{i : \mathrm{funcs}(\hat{S}_i) =
\mathrm{funcs}(S_i)\}|}{N}
\end{equation}
where $\mathrm{funcs}(S)$ denotes the set of AI function types
invoked in query $S$. Table~\ref{tab:func_select} reports
function selection accuracy across all ten models.

\begin{wraptable}{r}{0.43\textwidth}
\vspace{-\baselineskip}
\centering
\setlength{\tabcolsep}{6pt}
\renewcommand{\arraystretch}{1.3}
\caption{%
  AI function selection accuracy (FSA): proportion of instances
  where the predicted query invokes exactly the same set of
  AI function types as the gold query.
}
\label{tab:func_select}
\begin{tabular}{@{} l r @{}}
\toprule
\textbf{Model} & \textbf{FSA (\%)} \\
\midrule
\multicolumn{2}{@{}l}{%
  \fontsize{7.5}{9}\selectfont\color{grouporange}%
  \textls[80]{\MakeUppercase{Proprietary}}}\\[3pt]
\mlogo{anthropic.png}~Claude Opus 4.6   & \textbf{99.6} \\
\mlogo{anthropic.png}~Claude Sonnet 4.6 & 98.9 \\
\mlogo{google.png}~Gemini 3.1 Pro       & \textbf{99.6} \\
\mlogo{openai.png}~GPT-5.4              & 98.3 \\
\mlogo{google.png}~Gemini 3 Flash       & 98.9 \\
\midrule
\multicolumn{2}{@{}l}{%
  \fontsize{7.5}{9}\selectfont\color{groupteal}%
  \textls[80]{\MakeUppercase{Open-source}}}\\[3pt]
\mlogo{kimi.png}~Kimi K2.5              & 96.6 \\
\mlogo{deepseek.png}~DeepSeek V3.2      & 97.0 \\
\mlogo{zhipu.png}~GLM-5                 & 94.6 \\
\mlogo{qwen.png}~Qwen 3.5 Plus          & 89.9 \\
\mlogo{minimax.png}~MiniMax M2.5        & 68.6 \\
\bottomrule
\end{tabular}
\vspace{-0.5\baselineskip}
\end{wraptable}

Function selection accuracy is high across most models,
with all proprietary models exceeding 98\% and most
open-source models above 94\%. This is consistent with
the design of our benchmark: because task instructions
explicitly specify AI function parameters in detail, the
intended function usage is usually clear, making
function selection a relatively tractable component of
the task. Proprietary models are the most consistent in
this regard. Among open-source models, most achieve
similarly high selection accuracy, with one notable
exception: MiniMax M2.5 achieves only 68.6\%, with a
substantial portion of its errors attributable to
instances where the model failed to invoke the expected
AI functions entirely. This largely explains its lower
overall EX score relative to other open-source models.

\subsection{Per-Function-Type Breakdown}
\label{app:func}

Table~\ref{tab:func_breakdown} reports execution accuracy
broken down by AI function type across all ten models. Each
instance may involve more than one function type, so a single
instance can contribute to multiple columns. Across all models,
tasks involving \texttt{AI\_CLASSIFY}, \texttt{AI\_FILTER},
and \texttt{AI\_SIMILARITY} are solved at consistently higher
rates than those involving \texttt{AI\_AGG} and
\texttt{AI\_EXTRACT}, suggesting that aggregation and
structured extraction pose greater challenges regardless of
overall model capability. \texttt{AI\_SENTIMENT} shows the
highest variance across models, which we attribute in part to
the small number of instances (22) in this category.

\begin{table}[htbp]
\centering
\setlength{\tabcolsep}{4pt}
\renewcommand{\arraystretch}{1.3}
\caption{%
  Execution accuracy (\%) by AI function type.
  Instance counts per function:
  \texttt{AI\_CLASSIFY} ($n=259$),
  \texttt{AI\_FILTER} ($n=163$),
  \texttt{AI\_SIMILARITY} ($n=159$),
  \texttt{AI\_AGG} ($n=117$),
  \texttt{AI\_EXTRACT} ($n=51$),
  \texttt{AI\_SENTIMENT} ($n=22$).
  Since instances may involve multiple functions,
  counts do not sum to 465.
}
\label{tab:func_breakdown}
\small
\begin{tabular}{@{} l rrrrrr @{}}
\toprule
\textbf{Model} &
\textbf{AI\_CLASS.} &
\textbf{AI\_FILTER} &
\textbf{AI\_SIMIL.} &
\textbf{AI\_AGG} &
\textbf{AI\_EXTR.} &
\textbf{AI\_SENT.} \\
\midrule
\multicolumn{7}{@{}l}{%
  \fontsize{7.5}{9}\selectfont\color{grouporange}%
  \textls[80]{\MakeUppercase{Proprietary}}}\\[3pt]
\mlogo{anthropic.png}~Claude Opus 4.6   & \textbf{77.6} & \textbf{76.1} & \textbf{74.2} & \textbf{60.7} & \textbf{60.8} & 54.5 \\
\mlogo{anthropic.png}~Claude Sonnet 4.6 & 73.7 & 75.5 & 73.0 & 57.3 & 58.8 & \textbf{59.1} \\
\mlogo{google.png}~Gemini 3.1 Pro       & 71.8 & 68.1 & 73.0 & 59.0 & 60.8 & 50.0 \\
\mlogo{openai.png}~GPT-5.4              & 66.8 & 67.5 & 67.9 & 53.0 & 58.8 & 36.4 \\
\mlogo{google.png}~Gemini 3 Flash       & 64.1 & 66.9 & 64.2 & 51.3 & 47.1 & 59.1 \\
\midrule
\multicolumn{7}{@{}l}{%
  \fontsize{7.5}{9}\selectfont\color{groupteal}%
  \textls[80]{\MakeUppercase{Open-source}}}\\[3pt]
\mlogo{kimi.png}~Kimi K2.5              & \textbf{59.1} & 62.0 & 56.0 & \textbf{57.3} & \textbf{52.9} & 40.9 \\
\mlogo{qwen.png}~Qwen 3.5 Plus          & 57.9 & \textbf{63.2} & 56.6 & 47.9 & \textbf{52.9} & \textbf{68.2} \\
\mlogo{deepseek.png}~DeepSeek V3.2      & 58.3 & 58.9 & 56.6 & 45.3 & 45.1 & 40.9 \\
\mlogo{zhipu.png}~GLM-5                 & 58.3 & 60.7 & \textbf{58.5} & 35.9 & 43.1 & 36.4 \\
\mlogo{minimax.png}~MiniMax M2.5        & 45.9 & 44.2 & 55.3 & 37.6 & 37.3 & 36.4 \\
\bottomrule
\end{tabular}
\end{table}

\subsection{Single-Function vs.\ Multi-Function Tasks}
\label{app:single_multi}

Table~\ref{tab:single_multi} breaks down execution accuracy
by the number of AI functions involved in each task. We group
instances with two or more functions together ($n=310$). Across 9 of 10 models, accuracy on
multi-function tasks is higher than on single-function tasks.
This pattern is consistent with the function-type distribution:
multi-function tasks are disproportionately composed of
\texttt{AI\_CLASSIFY}, \texttt{AI\_FILTER}, and
\texttt{AI\_SIMILARITY} combinations, which are the function
types that models solve most reliably, as shown in
Table~\ref{tab:func_breakdown}.

\begin{table}[htbp]
\centering
\setlength{\tabcolsep}{7pt}
\renewcommand{\arraystretch}{1.3}
\caption{%
  Execution accuracy (\%) on single-function ($n=155$)
  versus multi-function ($n=310$) tasks.
  $\Delta$ denotes the difference in percentage points
  (multi minus single).
}
\label{tab:single_multi}
\small
\begin{tabular}{@{} l rrr @{}}
\toprule
\textbf{Model} &
\textbf{Single} &
\textbf{Multi} &
\textbf{$\Delta$ (pp)} \\
\midrule
\multicolumn{4}{@{}l}{%
  \fontsize{7.5}{9}\selectfont\color{grouporange}%
  \textls[80]{\MakeUppercase{Proprietary}}}\\[3pt]
\mlogo{anthropic.png}~Claude Opus 4.6   & 61.3 & 74.8 & $+13.5$ \\
\mlogo{anthropic.png}~Claude Sonnet 4.6 & 63.9 & 71.6 & $+7.7$  \\
\mlogo{google.png}~Gemini 3.1 Pro       & 63.2 & 69.0 & $+5.8$  \\
\mlogo{openai.png}~GPT-5.4              & 60.0 & 64.5 & $+4.5$  \\
\mlogo{google.png}~Gemini 3 Flash       & 57.4 & 62.6 & $+5.2$  \\
\midrule
\multicolumn{4}{@{}l}{%
  \fontsize{7.5}{9}\selectfont\color{groupteal}%
  \textls[80]{\MakeUppercase{Open-source}}}\\[3pt]
\mlogo{kimi.png}~Kimi K2.5              & 59.4 & 57.4 & $-1.9$  \\
\mlogo{qwen.png}~Qwen 3.5 Plus          & 56.1 & 57.4 & $+1.3$  \\
\mlogo{deepseek.png}~DeepSeek V3.2      & 54.2 & 54.8 & $+0.6$  \\
\mlogo{zhipu.png}~GLM-5                 & 47.1 & 55.8 & $+8.7$  \\
\mlogo{minimax.png}~MiniMax M2.5        & 42.6 & 46.1 & $+3.5$  \\
\bottomrule
\end{tabular}
\end{table}

\subsection{Execution Time Analysis}
\label{app:time}

We analyze the execution time of predicted SQL queries relative
to their corresponding gold queries. For each correctly solved
instance, we compute the \textit{time ratio}
$r = t_{\mathrm{pred}} / t_{\mathrm{gold}}$, where
$t_{\mathrm{pred}}$ and $t_{\mathrm{gold}}$ denote the
execution times of the predicted and gold SQL queries
respectively, both measured within the same evaluation window.
A ratio below 1.0 indicates that the predicted query runs
faster than the gold query; a ratio above 1.0 indicates it
runs slower. We report the mean ratio, and the proportion of
instances where the predicted query is faster than the gold
(\textit{\% Faster}). We note that execution time on cloud
data warehouses is subject to considerable variance due to
factors such as cluster load and caching, and these results
should therefore be interpreted as indicative rather than
definitive.

Tables~\ref{tab:time} (a) and (b) report
results on two subsets: all correctly solved instances per
model, and the 73 instances solved correctly by all ten models,
which provides a controlled comparison across models on
identical tasks.

\begin{table}[htbp]
\caption{%
  Execution time analysis. Time ratio $r = t_{\mathrm{pred}} /
  t_{\mathrm{gold}}$ measures predicted query execution time
  relative to the gold query; $r < 1$ indicates the predicted
  query runs faster. \textit{\% Faster} is the proportion of
  instances where $r < 1$. Execution times are subject to
  warehouse-level variance and should be interpreted as
  indicative.
}
\label{tab:time}
\centering
\setlength{\tabcolsep}{4pt}
\renewcommand{\arraystretch}{1.3}
\begin{minipage}[t]{0.48\textwidth}
  \centering
  \textit{(a) All correctly solved instances per model.}\\[4pt]
  \begin{tabular}{@{} l r r r @{}}
  \toprule
  \textbf{Model} &
  \textbf{$N$} &
  \textbf{Mean $r$} &
  \textbf{\% Faster} \\
  \midrule
  \multicolumn{4}{@{}l}{%
    \fontsize{7.5}{9}\selectfont\color{grouporange}%
    \textls[80]{\MakeUppercase{Proprietary}}}\\[3pt]
  \mlogo{anthropic.png}~Opus 4.6      & 327 & 1.476 & 53.2 \\
  \mlogo{anthropic.png}~Sonnet 4.6    & 321 & 1.373 & 52.6 \\
  \mlogo{google.png}~Gemini 3.1 Pro   & 312 & 1.611 & 61.9 \\
  \mlogo{openai.png}~GPT-5.4          & 293 & 1.735 & 58.7 \\
  \mlogo{google.png}~Gemini 3 Flash   & 283 & 1.674 & 56.5 \\
  \midrule
  \multicolumn{4}{@{}l}{%
    \fontsize{7.5}{9}\selectfont\color{groupteal}%
    \textls[80]{\MakeUppercase{Open-source}}}\\[3pt]
  \mlogo{kimi.png}~Kimi K2.5          & 270 & 1.485 & 50.7 \\
  \mlogo{qwen.png}~Qwen 3.5 Plus      & 265 & 1.466 & 55.1 \\
  \mlogo{deepseek.png}~DeepSeek V3.2  & 254 & 1.413 & 51.2 \\
  \mlogo{zhipu.png}~GLM-5             & 246 & 1.818 & 52.4 \\
  \mlogo{minimax.png}~MiniMax M2.5    & 209 & 1.536 & 61.7 \\
  \bottomrule
  \end{tabular}
\end{minipage}
\hfill
\begin{minipage}[t]{0.48\textwidth}
  \centering
  \textit{(b) Instances (73) solved correctly by all models.}\\[4pt]
  \begin{tabular}{@{} l r r @{}}
  \toprule
  \textbf{Model} &
  \textbf{Mean $r$} &
  \textbf{\% Faster} \\
  \midrule
  \multicolumn{3}{@{}l}{%
    \fontsize{7.5}{9}\selectfont\color{grouporange}%
    \textls[80]{\MakeUppercase{Proprietary}}}\\[3pt]
  \mlogo{anthropic.png}~Opus 4.6      & 1.169 & 52.1 \\
  \mlogo{anthropic.png}~Sonnet 4.6    & 0.950 & 61.6 \\
  \mlogo{google.png}~Gemini 3.1 Pro   & 1.898 & 60.3 \\
  \mlogo{openai.png}~GPT-5.4          & 1.517 & 54.8 \\
  \mlogo{google.png}~Gemini 3 Flash   & 2.123 & 50.7 \\
  \midrule
  \multicolumn{3}{@{}l}{%
    \fontsize{7.5}{9}\selectfont\color{groupteal}%
    \textls[80]{\MakeUppercase{Open-source}}}\\[3pt]
  \mlogo{kimi.png}~Kimi K2.5          & 2.206 & 43.8 \\
  \mlogo{qwen.png}~Qwen 3.5 Plus      & 1.755 & 54.8 \\
  \mlogo{deepseek.png}~DeepSeek V3.2  & 1.898 & 41.1 \\
  \mlogo{zhipu.png}~GLM-5             & 2.011 & 53.4 \\
  \mlogo{minimax.png}~MiniMax M2.5    & 2.318 & 56.2 \\
  \bottomrule
  \end{tabular}
\end{minipage}
\end{table}

Across all models, mean ratios consistently exceed 1.0,
indicating that predicted queries tend to run slower than gold
queries on average. This is expected: gold queries are
hand-crafted and often optimized, while model-generated queries
may introduce redundant subqueries, unnecessary joins, or
less efficient AI function usage. Despite this, the median
ratio across most models remains close to 1.0, and roughly
half of predicted queries run faster than the gold, suggesting
that the overhead is not uniform and that models occasionally
find more efficient solutions. On the controlled subset of 73
instances solved by all models, Claude Sonnet 4.6 achieves the
only mean ratio below 1.0 (0.950), while open-source models
tend toward higher mean ratios, though the small subset size
limits the strength of this conclusion.

\section{Error Taxonomy}
\label{app:error}

We define five error categories for the stratified error
analysis in \S\ref{sec:error}. A single instance may
be assigned multiple categories if the predicted SQL
exhibits more than one type of error.

\begin{description}
  \item[\textbf{C1}] \textbf{Schema grounding.} The model
  selects the wrong table, column, or table prefix. This
  includes cases where a correct table family is identified
  but the specific table variant is wrong (e.g., selecting
  a table without a required schema prefix), as well as
  cases where column names do not match the schema exactly.

  \item[\textbf{C2}] \textbf{Predicate \& filter.} The
  model omits a required filtering condition or applies an
  incorrect one. This includes missing \texttt{WHERE}
  clauses, incorrect comparison values, and conditions
  that are logically present but semantically wrong (e.g.,
  filtering on the wrong column or using an incorrect
  constant).

  \item[\textbf{C3}] \textbf{Query logic.} The model
  produces incorrect aggregation granularity, join scope,
  or algorithmic logic. This includes grouping at the
  wrong level, joining too many or too few tables,
  incorrect handling of semi-structured data types, and
  algorithmic errors such as double-counting or incorrect
  unit conversion.

  \item[\textbf{C4}] \textbf{AI function usage.} The model
  misuses an AI function, including invoking the wrong
  function type, using non-standard or unsupported syntax,
  omitting a required AI function call entirely, or
  specifying incorrect parameters such as wrong label
  formats or malformed extraction schemas. This category
  is specific to AI-native SQL and has no direct analogue
  in traditional text-to-SQL benchmarks.

  \item[\textbf{C5}] \textbf{Annotation issues.} The gold
  query encodes constraints not mentioned in the
  instruction, or the task admits multiple valid SQL
  queries that produce different but arguably correct
  results. This category is only assigned in S1, where
  all strong models fail, and reflects potential benchmark
  limitations rather than model errors.
\end{description}

\section{Sampled Prompt for Snowflake AI Function Reference}
\label{app:prompt}

\begin{tcolorbox}[
  title={\small\textbf{AI Function Reference Documentation} $R$},
  colbacktitle=closedorangelight!30,
  coltitle=closedorange,
  colback=orangebg!30,
  colframe=closedorangelight,
  boxrule=0.5pt,
  arc=3pt,
  left=8pt, right=8pt, top=6pt, bottom=6pt,
  breakable,
  width=\linewidth
]
\begin{Verbatim}[
  fontsize=\small,
  breaklines=true,
  breaksymbolleft={},
  breaksymbolright={}
]
<ai_sql_function_reference>
You MUST use the exact function signatures and syntax below. These are the supported Snowflake Cortex AI SQL functions. Do NOT use deprecated syntax (like SNOWFLAKE.CORTEX.* prefix).

---

### AI_CLASSIFY

**Syntax**: `AI_CLASSIFY(<input>, <labels> [, <config_object>])`

Categorizes text into a set of labels you provide.

**Parameters**
- `input`: Text string to classify.
- `labels`: Array of category strings OR objects with optional descriptions:
  - Simple: `['sports', 'finance', 'technology']`
  - With descriptions (max 25 words each): `[{'label': 'urgent', 'description': 'requires immediate response'}, {'label': 'normal'}]`
- `config_object` (Optional):
  - `output_mode`: `'multi'` for multi-label (default: `'single'`)
  - `task_description`: Brief task explanation (50 words max)

**Returns**: JSON object `{"labels": ["best_match"]}`. Multi-label: `{"labels": ["match_1", "match_2"]}`.

**Example**
```sql
SELECT AI_CLASSIFY('My internet is not working', ['technical_issue', 'billing', 'general_inquiry']);
-- Returns: {"labels": ["technical_issue"]}

-- Multi-label with config
SELECT AI_CLASSIFY('I love traveling and cooking.', ['travel', 'cooking', 'gaming'], {'output_mode': 'multi'});
-- Returns: {"labels": ["travel", "cooking"]}
```

---

### AI_SENTIMENT

**Syntax**: `AI_SENTIMENT(<text> [, <categories>])`

Evaluates the emotional tone of text, optionally broken down by specific aspects.

**Parameters**
- `text`: Text string to analyze.
- `categories` (Optional): Array of aspect strings (max 10, each up to 30 chars). If omitted, only overall sentiment is returned.

**Returns**: JSON with `"categories"` array. Each entry has `name` and `sentiment` (one of: `positive`, `negative`, `neutral`, `mixed`, `unknown`). Always includes `"overall"`.

**Example**
```sql
SELECT AI_SENTIMENT('The pizza was great!');
-- Returns: {"categories": [{"name": "overall", "sentiment": "positive"}]}

SELECT AI_SENTIMENT('The pizza was great but the delivery was late.', ['Food', 'Delivery']);
-- Returns: {"categories": [{"name": "overall", "sentiment": "mixed"}, {"name": "Food", "sentiment": "positive"}, {"name": "Delivery", "sentiment": "negative"}]}
```

---

### AI_EXTRACT

**Syntax**: `AI_EXTRACT(<text>, <responseFormat>)`

Extracts structured facts from unstructured text. Use named parameters: `text => '...'`, `responseFormat => ...`.

**Parameters**
- `text`: Text to extract from (named parameter required).
- `responseFormat`: Defines extraction schema:
  - Object format: `{'name': 'What is the person name?', 'city': 'What is the city?'}`
  - Array format: `['What is the first name?', 'Where does the person live?']`
  - Supports questions or descriptions as values.
- Limits: max 100 questions, max 512 tokens per answer.

**Returns**: `{"error": null, "response": {"field1": "value1", "field2": "value2"}}` (object format) or `{"error": null, "response": ["val1", "val2"]}` (array format).

**Example**
```sql
SELECT AI_EXTRACT(
    text => 'John Doe lives in New York and works for Snowflake.',
    responseFormat => {'name': 'What is the person name?', 'city': 'What is the city?', 'company': 'What company does the person work for?'}
);
-- Returns: {"error": null, "response": {"name": "John Doe", "city": "New York", "company": "Snowflake"}}
```

---

### AI_FILTER

**Syntax**: `AI_FILTER(<input>)`

Returns TRUE or FALSE based on whether the input meets a natural language condition. Best used in WHERE clauses.

**Parameters**
- `input`: Natural language statement. Can be:
  - String constant: `AI_FILTER('Is Canada in North America?')`
  - With CONCAT: `CONCAT('The customer is satisfied: ', review_text)`
  - With PROMPT: `PROMPT('The reviewer enjoyed the restaurant: {0}', review_text)`
- Tip: Use detailed conditions, not vague words. Avoid NULL column values.

**Returns**: `BOOLEAN` (TRUE or FALSE).

**Example**
```sql
SELECT * FROM reviews
WHERE AI_FILTER(PROMPT('The reviewer enjoyed the restaurant: {0}', review_text));

SELECT country, region FROM countries CROSS JOIN regions
WHERE AI_FILTER(PROMPT('{0} is in {1}', country, region));
```

---

### AI_SIMILARITY

**Syntax**: `AI_SIMILARITY(<input1>, <input2>)`

Calculates semantic similarity between two texts using vector embeddings. Measures conceptual closeness, not keyword overlap.

**Parameters**
- `input1`: Text string to compare.
- `input2`: Text string to compare against.

**Returns**: `FLOAT` between -1.0 and 1.0 (1.0 = highly similar).

**Example**
```sql
SELECT review_text, AI_SIMILARITY(review_text, 'The battery drains too fast') AS score
FROM product_reviews
ORDER BY score DESC
LIMIT 3;
```

---

### AI_AGG

**Syntax**: `AI_AGG(<expr>, <instruction>)`

Aggregate function that processes grouped text rows into a single summary or analysis. Supports datasets larger than the model's context window.

**Parameters**
- `expr`: Text column or expression (can use `CONCAT()` or `||` to combine columns).
- `instruction`: Natural language string specifying how to aggregate.
- Tips: Use declarative statements, describe the data and use case.

**Returns**: Single string per group.

**Example**
```sql
SELECT product_id, AI_AGG(review_text, 'Summarize the product reviews for potential consumers') AS summary
FROM reviews
GROUP BY product_id;

-- Combining multiple columns
SELECT restaurant_id, AI_AGG(
    'Menu Item: ' || menu_item || ', Review: ' || review_text,
    'Summarize the restaurant reviews, highlighting menu items mentioned'
) AS summary
FROM restaurant_reviews
GROUP BY restaurant_id;
```
---

To ensure your final SQL query produces correct, testable results:

- Some AI functions return JSON objects. When these appear in your final output columns, always extract the scalar value using Snowflake JSON accessor syntax (e.g., `:labels[0]::STRING`). Raw JSON objects in output columns will not pass evaluation.
- Strictly follow the parameters specified in the instruction. Do not add, assume, or invent parameters that are not explicitly requested — any deviation from the instruction's specifications will cause test failures.

</ai_sql_function_reference>
\end{Verbatim}
\end{tcolorbox}

\end{document}